\documentclass{article}
\usepackage{arxiv}

\usepackage[english]{babel}

\usepackage[utf8]{inputenc}
\usepackage{bm}

\usepackage{xspace}
\usepackage{amsmath,mathtools}
\usepackage{dsfont,bbm}

\usepackage{bbm}
\usepackage{url}
\usepackage[caption=false]{subfig}
\usepackage[noend]{algpseudocode}
\usepackage[vlined,linesnumbered,ruled,titlenotnumbered]
{algorithm2e}
\usepackage{xcolor}
\usepackage{tikz}
\usepackage{graphicx}
\renewcommand{\labelenumi}{(\alph{enumi})}
\renewcommand\theenumi\labelenumi
\usepackage{diagbox}
\usepackage{tabularx}
\newcolumntype{Y}{>{\centering\arraybackslash}X}

\usepackage{amsfonts}

\newcommand{\ea}{\ooea}

\newcommand{\ooea}{(1+1)~EA\xspace}

\newcommand{\om}{\textsc{OneMax}\xspace}

\newcommand{\R}{\ensuremath{\mathbb{R}}}

\DeclareMathOperator{\Prob}{Pr}

\newcommand{\smin}{s_{\mathrm{min}}}
\newcommand{\smax}{s_{\mathrm{max}}}

\newcommand{\ones}[1]{\lvert #1\rvert_1}

\newcommand{\ie}{i.\,e.\xspace}

\newcommand{\wlo}{w.\,l.\,o.\,g.\xspace}

\newcommand{\card}[1]{\lvert #1\rvert}
\usepackage{amsmath}
\usepackage{amsthm, amssymb}
\usepackage{mathtools}

\newcommand{\prob}[1]{\mathord{\Prob}\mathord{\left(#1\right)}}

\newcommand{\expect}[1]{\mathord{\mathrm{E}}\mathord{\left(#1\right)}}

\newtheorem{definition}{Definition}

\newtheorem{theorem}{Theorem}

\newenvironment{proofof}[1]{\begin{proof}[Proof of~#1]}{\end{proof}}

\algnewcommand{\IfThenElse}[3]{% \IfThenElse{<if>}{<then>}{<else>}
  \State \algorithmicif\ #1\ \algorithmicthen\ #2\ \algorithmicelse\ #3}

\algnewcommand{\IfThen}[2]{% \IfThenElse{<if>}{<then>}{<else>}
  \State \algorithmicif\ #1\ \algorithmicthen\ #2}

\title{Runtime Analysis of the (1+1)~EA on Weighted Sums of Transformed Linear Functions}
\author{Frank Neumann\\
Optimisation and Logistics\\
School of Computer Science\\
The University of Adelaide\\
Adelaide, Australia
\And
Carsten Witt\\
Algorithms, Logic and Graphs\\
DTU Compute\\ Technical University of Denmark\\
2800 Kgs. Lyngby Denmark
}

\begin{document}

\maketitle
\begin{abstract}
  Linear functions play a key role in the runtime analysis of evolutionary algorithms and studies have provided a wide range of new insights and techniques for analyzing evolutionary computation methods. Motivated by studies on separable functions and the optimization behaviour of evolutionary algorithms as well as objective functions from the area of chance constrained optimization, we study the class of objective functions that are weighted sums of two transformed linear functions. Our results show that the (1+1)~EA, with a mutation rate depending on the number of overlapping bits of the functions, obtains an optimal solution for these functions in expected time $O(n \log n)$, thereby generalizing a well-known result for linear functions to a much wider range of problems.
\end{abstract}

\section{Introduction}
Runtime analysis is one of the major theoretical tools to provide rigorous insights into the working behavior of evolutionary algorithms and other randomized search heuristics~\cite{NeumannW10,Jansen13,DoerrN20}.
The class of pseudo-Boolean linear functions plays a key role in the area of runtime analysis. Starting with the simplest linear functions called OneMax for which the first runtime analysis has been carried out, a wide range of results have been obtained for the general class of linear functions. %\cite{djwea02, He2004, JJ08, Jagerskupper11, DJWLinearRevisited, WittCPC13}. 
This includes the study of Droste, Jansen and Wegener~\cite{djwea02}
who were the first to obtain
an upper bound of $O(n\log n)$ for the \ea on the general class of pseudo-Boolean linear functions. This groundbreaking result has been based on a very lengthy proof and subsequently a wide range of improvements have been made in terms of the development of new techniques for the analysis as well as the precision of the results.
%$O(n \log n)$ as well as several indepth studies improving proof techniques and analyzing the leading constants.
%One of the classes of problems which has been studied for a simple %evolutionary algorithm, called \ea, is the 
% class of linear pseudo-boolean functions~\cite{djwea02, He2004, JJ08, %Jagerskupper11, DJWLinearRevisited, WittCPC13}. 
%The problem is to optimise a linear function of $n$ Boolean variables.  %Droste, Jansen and Wegener \cite{djwea02}
% were the first to obtain
%an upper bound of $O(n\log n)$ 
%for the expected optimisation time of the \ea on this problem, where 
%the presented proof is highly technical. 
The proof has been simplified significantly  using the analytic framework of drift analysis~\cite{Hajek1982} by  
He and Yao~\cite{He2004}. 
%presented a simplified proof for the same upper bound of $O(n \log n)$~\cite{ He2004}.
%Another 
%major improvement was made in~
J{\"a}gersk{\"u}pper~\cite{JJ08,Jagerskupper11} provided the first  analysis of the leading coefficient in 
the bound $O(n\log n)$ on the 
 the optimisation time for the problem. 
Furthermore, advances to simplify proofs and getting precise results have been made using the
framework of multiplicative 
drift~\cite{DJWMultiplicativeDrift}.
Doerr, Johannsen and Winzen  improved the 
upper bound result 
to  $(1.39 + o(1))en \ln n$~\cite{DJWLinearRevisited}.  Finally, Witt~\cite{WittCPC13} 
improved this bound  to $en \ln n + O(n)$ by using adaptive 
drift analysis~\cite{DoerrGoldbergLinear,DoerrGoldbergAdaptive}. We expand such investigations for the \ea into a wider class of problems that are modelled by two transformed linear functions. This includes classes of separable functions and chance constrained optimization problems.
%by making use of a novel potential function.

\subsection{Separable Functions}

As an example, consider the separable objective function
\begin{equation}
\label{eq:sep}
f(x) =
\left(\sum_{i=1}^{n/2} w_i x_i \right)^2 + \sqrt{\sum_{i=n/2+1}^n w_i x_i}
\end{equation}
where $w_i \in \mathds{Z}^+$, $1 \leq i \leq n$, and  $x = (x_1, \dots, x_n) \in \{0,1\}^n$. The function $f$ consists of two objective functions
$$f_1(x_1, \ldots, x_{n/2}) = \left(\sum_{i=1}^{n/2} w_i x_i\right)^2 \text{ and }  f_2(x_{n/2+1}, \ldots, x_n)=\sqrt{\sum_{i=n/2+1}^n w_i x_i}.$$ 
%\frank{Better do $f_1$ as square of linear function?? Shows that both can be non-linear.} \carsten{Yes, would be good %to 
%include square or another non-linear transformation.}

Here $f_1$ is the square of a function linear in the first half of variables and $f_2$ is the square root of a linear function in the remaining variables. Some investigations on how evolutionary algorithms optimize separable fitness functions have been carried out in~\cite{DBLP:conf/foga/DoerrSW13}. 
It has been shown that if the different functions only have a small range, then the \ea optimizes separable functions efficiently if the different separable functions themselves are easy to be optimized. However, in our example above the two separable functions may take on exponentially many values but both functions on their own are optimized by the \ea in time $O(n \log n)$ using the results for the \ea on linear functions. This holds as the transformation applying the square in $f_1$ or the square root in $f_2$ does not change the behavior of the \ea. The questions arises whether the $O(n \log n)$ bounds also holds for the function $f$ which combines~$f_1$ and $f_2$. We investigate this setting of separable functions for the more general case where the objective function is given as a weighted sum of two separable transformed linear functions. For technical 
reasons, we consider a \ooea with potentially reduced mutation probability depending on the number of overlapping bits of the
two functions.

\subsection{Chance Constrained Problems}
Another motivation for our work comes from problems from the area of chance constrained optimization~\cite{Charnes} and considers the case where the two functions are overlapping or are even defined on the same set of variables.
Recently evolutionary algorithms have been used for chance constrained problems which motivates our investigations. In a chance constrained setting the input involves stochastic components and the goal is to optimize a given objective function under the condition that constraints are met with high probability or that function values are guaranteed with a high probability. Evolutionary algorithms have been designed  for the chance constrained knapsack problem~\cite{DBLP:conf/gecco/XieHAN019,DBLP:conf/gecco/XieN020,DBLP:conf/ecai/AssimiHXN020}, chance constrained stock pile blending problems~\cite{DBLP:conf/gecco/XieN021}, and chance constrained submodular functions~\cite{DBLP:conf/ppsn/NeumannN20}.

Runtime analysis results have been obtained for restricted settings of the knapsack problem~\cite{DBLP:conf/foga/0001S19,DBLP:conf/gecco/XieN0S21} where the weights are stochastic and the constraint bound has to be met with high probability.
The analysis for the case of stochastic constraints and the class of submodular function~\cite{DBLP:conf/aaai/DoerrD0NS20} and the knapsack problem~\cite{DBLP:conf/gecco/XieHAN019} already reveal constraint functions that are a linear combination of the expected weight and the  standard deviation of a solution when using Chebyshev's inequality for constraint evaluation. Such functions are the subject of our investigations.

To make the type of problems that we are interested in clear, we state the following problem. 
Given a set of $m$ items $E=\{e_1, \ldots, e_m\}$ with random weights $w_i$, $1 \leq i \leq m$. We assume that the weights are independent and each $w_i$ is distributed according to a normal distribution $N(\mu_i, \sigma_i^2)$, $1 \leq i \leq m$.  We assume $\mu_i \geq 0$ and $\sigma_i\geq 0$, $1 \leq i \leq m$. 
Our goal is to 
\begin{equation}
\min W  \text{~~~~subject to~~~~}  \Prob( w(x) \leq W) \geq \alpha
\label{chance-problem}
\end{equation}
where $w(x) = \sum_{i=1}^n w_i x_i$, $x \in \{0,1\}^m$, and $\alpha \in \mathopen{]}0,1\mathclose{[}$. The problem given in Equation~\eqref{chance-problem} is usually considered under additional constraints, e.g. spanning tree constraints in~\cite{DBLP:journals/dam/IshiiSNN81}, which we do not consider in this paper.

According to \cite{DBLP:journals/dam/IshiiSNN81} the problem given in Equation~\ref{chance-problem} is equivalent to minimizing the fitness function
\begin{equation}
\label{eq:sumofidentityandsquare}
    g(x) = \sum_{i=1}^m \mu_i x_i + K_{\alpha} \left(\sum_{i=1}^m \sigma_i^2 x_i \right)^{1/2}
\end{equation}
where $K_{\alpha}$ is the $\alpha$-fractile point of the standard normal distribution.

The fitness function $g$ is a linear combination of the expected value of a solution which is a linear function and the square root of its variance where the variance is again a linear function. 
In order to understand the behaviour of evolutionary algorithms on fitness functions obtained for chance constrained optimization problems, our runtime analysis for the \ea covers such fitness functions if we assume the  reduced 
mutation probability mentioned above.

\subsection{Transformed Linear Functions}

In our investigations, we consider the much wider class of problems where a given fitness function is obtained by the linear combination of two transformed linear functions. 
The transformations applied to the linear functions only have to be monotonically increasing in terms of the functions values of the linear functions.
This includes the setting of separable functions and chance constrained problems described previously. Furthermore, we do not require that the two linear functions are defined on the same number of bits.

The main result of our paper is an $O(n \log n)$ upper bound for the \ea with mutation probability $1/(n+s)$ on the class of sums of two transformed linear functions where $s$ is the number of bits for which the two linear functions overlap. This directly transfers to the separable problem type given in Equation~\ref{eq:sep} with standard bit mutation probability $1/n$ and to the chance constraint formulation given in Equation~\ref{eq:sumofidentityandsquare} when using mutation probability $1/(2n)$.
  %and are independent of each other in the sense that they do not share the bits. 
%\frank{Does the previous sentence still apply/make sense?}\carsten{Shortened the sentence}

The outline of the paper is as follows. In Section~\ref{sec2}, we formally introduce the problem formulation for which we analyze the \ea in this paper. We discuss the exclusion of negative weights in our setup in Section~\ref{sec3} and present the $O(n \log n)$ bound in Section~\ref{sec4}. Finally, we finish with some discussion and conclusions.

\section{Preliminaries}
\label{sec2}

%\section{Analysis of \ea}
%\label{sec3}
The \ea shown in Algorithm~\ref{alg:EA} (generalized with a parameter~$s$ discussed below; classically~$s=0$ is assumed) 
is a simple evolutionary algorithm using independent bit flips and elitist selection. It is very well studied in the theory of evolutionary computation \cite{DoerrProbabilisticTools} and serves as a stepping stone towards the analysis of more complicated evolutionary algorithms. As common, in the area of runtime analysis, we measure the run time of the \ea by the number of iterations of the repeat loop. The optimization time refers to the number of fitness evaluations until an optimal solution has been obtained for the first time, and the expected optimization time refers to the expectation of this value.

\subsection{Sums of Two Transformed Linear Functions without Constraints}
\label{sec:two}
We will study the \ea on the scenario given in \eqref{eq:sep} and  \eqref{eq:sumofidentityandsquare}, assuming no additional constraints. In fact, we will generalize the scenario to the sum of two transformed pseudo-Boolean linear functions 
which may be (partially) overlapping. Note, that in \eqref{eq:sep} there is no overlap on the domains of the two linear functions and the transformations are the square and the square root, whereas in \eqref{eq:sumofidentityandsquare} there is complete overlap on the domains  and the transformations are the identity function and the square root.

The crucial observation in our analysis is that the scenario considered here extends the linear function problem \cite{WittCPC13} that is heavily investigated in the theory of evolutionary algorithms. Despite the simple structure of the problem, there is no clear fitness-distance correlation in the linear function problem, which makes the analysis of the global search 
operator of the  \ea difficult. 
If only local mutations are used, 
leading to the well known randomized local search (RLS) algorithm \cite{DoerrDoerr16}, then both the linear function problem and the generalized scenario considered here are very easy to analyze using standard 
coupon collector arguments \cite{MotwaniRaghavan}, leading 
to $O(n\log n)$ expected optimization time. For the globally searching \ea, we will obtain the same bound, proving that the problem is easy to solve for it; however, we need advanced drift analysis methods to prove this. 

We note that the class 
of functions we consider 
falls within the more 
general class of so-called monotone 
functions. Such functions 
can be difficult to 
optimize with a \ea using 
 mutation probabilities
larger than~$1/n$ \cite{DoerrJSWZ13}; however, 
it is also known that 
the standard \ea with 
mutation probability~$1/n$
as considered here 
optimizes all monotone 
functions in expected 
time 
$O(n \log^2 n)$  \cite{DBLP:conf/analco/LenglerMS19}. Our 
bound is by an asymptotic 
factor of $\log n$ better if $s=o(n)$. However, it should be noted 
that for $s=\Omega(n)$, the bound $O(n\log n)$ already follows
directly from~\cite{DoerrJSWZ13} since it corresponds to 
a mutation probability of $c/n$ for a constant~$c<1$. In fact,  
the fitness function~$g(x)$ arising from the chance-constrained 
scenario presented in~\eqref{eq:sumofidentityandsquare} above would 
fall into the case $s=n/2$.

%\carsten{Ich habe die Fitnessfunktion auf einem Bitstring der Laenge 
%$n-s$ definiert und die Mutationswkt. auf $1/n$ gesetzt. Ist natuerlich analog zu 
%einem Bitstring der Laenge $n$ und Mutationswkt. $1/(n+s)$. In den Rechnungen ist die 
%erste Sichtweise einfacher zu praesentieren}

\begin{algorithm}[t]
	 Choose $x \in \{0,1\}^{n-s}$ uniformly at random\;
\Repeat{$\mathit{stop}$}{
Create $y$ by flipping each bit $x_{i}$ of $x$ with probability $p=\frac{1}{n}$\; 
\If{$f(y) \leq f(x)$} {
  $x \leftarrow y;$}
    }
\caption{\ea for minimization of a pseudo-Boolean function $f\colon\{0,1\}^{n-s}\to \R$, where 
$s\in\{0,\dots,n/2\}$} \label{alg:EA} 
\end{algorithm}

\paragraph{Set-up.}  We will investigate a general optimization 
scenario involving two linear pseudo-Boolean functions in an unconstrained search space. 
The objective function is an arbitrarily weighted sum 
of monotone transformations of two linear functions defined on 
(possibly overlapping) subspaces of $\{0,1\}^{n-s}$ 
for some $s\ge 0$, where 
$s$ denotes the number of shared bits. Note that the introduction of this paper 
mentions a search space of dimension $n$ and a mutation probability of $p=1/(n+s)$ for 
the \ea. While the former perspective is more natural to present, 
from now on, we consider the asymptotically equivalent setting of search space dimension 
$n-s$ and mutation probability~$p=1/n$, which eases notation in the upcoming 
calculations.

Let $\alpha$ be a constant such 
that $1/2\le \alpha\le \ln (2-\epsilon)\approx 0.693-\epsilon/2$ for some constant~$\epsilon>0$ and 
assume that $\alpha n$ is an integer. 
We allow the subfunctions to depend on a number of bits in $[(1-\alpha)n,\alpha n$], including 
the balanced case that both subfunctions depend on exactly $n/2$ bits.
Formally,
we have
{\sloppy
\begin{itemize}
    \item     
     linear functions 
     $\ell_1\colon \{0,1\}^{\alpha n}\to \R \text{ and } 
     \ell_2\colon \{0,1\}^{(1-\alpha) n}\to \R,$ 
     where $\ell_1(y_1,\dots,y_{\alpha n}) = \sum_{i=1}^{\alpha n} w^{(1)}_i y_i$, and 
     similarly 
$\ell_2(z_1,\dots,z_{(1-\alpha) n}) = \sum_{i=1}^{(1-\alpha)n} w^{(2)}_i z_i$  with non-negative weights $w^{(1)}_i$ and  $w^{(2)}_i$.
  \item   $B_1\subseteq\{1,\dots,n\}$ and $B_2\subseteq\{1,\dots,n\}$, denoting the bit positions that $\ell_1$ resp.\ $\ell_2$ are defined on in the actual objective function $f\colon\{0,1\}^{n-s}\to \R$.
  \item The overlap count $s\coloneqq \card{B_1\cap B_2}$, where $s\le \min\{(1-\alpha)n,\alpha n\} = (1-\alpha) n \le n/2$
  \item the  linear functions with extended domain 
    $\ell^*_1(x_1,\dots,x_{n-s}) = \sum_{i\in B_1} w^{(1)}_{r^{(1)}(i)} x_i$ where $r^{(1)}(i)$ is the rank of~$i$ in~$B_1$ (with the smallest number  
    receiving rank number~$1$); and analogously  $\ell^*_2(x_1,\dots,x_{n-s}) = \sum_{i\in B_2} w^{(2)}_{r^{(2)}(i)} x_i$; note that  $\ell_1^*$ and $\ell_2^*$ only  depend essentially on $\alpha n$ and $(1-\alpha) n$ bits, respectively.
  \item monotone increasing functions 
$h_1\colon \R\to \R$ and $h_2\colon \R\to \R$.
\end{itemize}}
Then the objective function $f\colon\{0,1\}^{n-s}\to \R$, which \wlo\ is 
to be minimized, is given by
$$
f(x_1,\dots,x_{n-s}) = h_1(\ell^*_1(x_1,\dots,x_{n-s}))
 + h_2(\ell^*_2(x_1,\dots,x_{n-s})).
$$

For $s=0$, $h_1$ being the square function, and $h_2$ being the square root function, this matches the setting of separable functions given in Equation~\ref{eq:sep}.
This set-up also includes the case that 
\[
f(x_1,\dots,x_{m}) = \ell_{1}(x_1,\dots,x_{m}) + R\sqrt{\ell_{2}(x_1,\dots,x_{m})}
\]
for two $m$-dimensional, 
completely overlapping linear functions $\ell_1$ and $\ell_2$ and 
an arbitrary factor $R\ge 0$, as motivated and given in Equation~\ref{eq:sumofidentityandsquare}. Note that this matches our set-up with $n=2m$ and $s=n$.
%presented above in \eqref{eq:sumofidentityandsquare}.

For our analysis we will make use of the multiplicative drift theorem (Theorem~\ref{theo:multdrift-upper}) that has been introduced in 
\cite{DJWMultiplicativeAlgorithmica} and was enhanced with tail bounds by
 \cite{DoerrGoldbergAdaptive}. We use a 
 slightly generalised presentation that can be found in 
\cite{LehreW21}. 
 
%
%The adaptation of the multiplicative drift theorem 
%to arbitrary positive $\smin$-values 
%has first been stated %in~\citet{DJWMultiplicativeAlgorithmica}.

\begin{theorem} [Multiplicative Drift, cf. \cite{DJWMultiplicativeAlgorithmica,DoerrGoldbergAdaptive,LehreW21}] 
\label{theo:multdrift-upper}
Let $(X_t)_{t\ge 0}$, be a stochastic process, adapted to a filtration~$\mathcal{F}_t$, 
over some state space $S\subseteq \{0\}\cup [\smin,\smax]$, where $0\in S$ and 
$\smin>0$. Suppose that there exists a $\delta>0$ such that
for all $t\ge 0$
\[
\expect{X_t-X_{t+1}\mid \mathcal{F}_t}\ge \delta X_t.
\]
\parbox{\textwidth}{%
Then it holds for the first hitting time 
$T:=\min\{t\mid X_t=0\}$ that 
\[
\expect{T\mid \mathcal{F}_0} \le \frac{\ln(X_0/\smin)+1}{\delta}.
\]
Moreover, 
$\Prob(T> (\ln(X_0/\smin)+r)/\delta) \le e^{-r} $ for any $r>0$.}
\end{theorem}

\section{Negative Weights Allow for Multimodal Functions}
\label{sec3}

We will now justify that the inclusion of negative weights in the underlying linear functions, along with 
overlapping domains, can lead to 
multimodal problems that cannot be optimized in expected time $O(n\log n)$ any longer. In the following example, 
the two linear functions depend essentially on all~$n$ bits.

Let 
\[
f(x_1,\dots,x_n) = \underbrace{\left(\frac{x_1}{2} + \sum_{i=2}^n  x_i\right)}_{h_1(\ell_1(x))} + \underbrace{ \left(\frac{\sum_{i=1}^n (1-x_i)}{n-0.5}\right)^{n^2}}_{h_2(\ell_2(x))}
\]
Basically, the first linear function $\ell_1(x)=x_1/2 + \sum_{i=2}^n  x_i$ is a \om function except for the first bit 
that has a smaller weight than the rest. The second linear function $\ell_2(x)$ is linear in the number of zeros, \ie, corresponds 
to the \textsc{ZeroMax} function that is equivalent to \om for the \ea due to symmetry reasons. The transformation $h_2$ 
that is applied to \textsc{ZeroMax} divides the number of zero-bits by $n-0.5$ and raises the result to a large power.
Essentially, the value of $h_2(z)$ is $e^{\Theta(n)}$ if $z=n$ and $e^{-\Theta(n)}$ otherwise. This puts a constraint 
on the number of zero-bits. If $\ones{x}\ge 1$, then $f$ is monotone increasing in $\ell_1$, \ie, search 
points decreasing the $\ell_1$-value also decrease the $f$-value. However, the all-zeros string has the largest~$f$-value, \ie, 
is worst.

We can now see that all search points having one exactly one one-bit at one of the positions $2,\dots,n$ are local optima. 
To create the global optimum from such a point, two bits have to flip simultaneously, leading to $\Omega(n^2)$ expected time 
to reach the optimum from the point. The situation is similar to the optimization of linear function under uniform constraints 
\cite{NeumannPW21}.

% This puts a constraint 
% on the number of zero-bits. If $\ones{x}\ge 1$, then $f$ is monotone increasing in $\ell_1$, \ie, search 
% points decreasing the $\ell_1$-value also decrease the $f$-value. However, the all-zeros string has the largest~$f$-value, \ie, 
% is worst.

% We can now see that all search points having exactly one one-bit at one of the positions $2,\dots,n$ are local optima. 
% To create the global optimum from such a point, two bits have to flip simultaneously, leading to $\Omega(n^2)$ expected time 
% to reach the optimum from the point. The situation is similar to the optimization of linear function under uniform constraints 
% \cite{NeumannPW21}.

% The assumption of all weights being positive is crucial. If we allowed the underlying linear functions to have negative weights, we could use a combination of a linear function and 
% \textsc{ZeroMax} along with a non-linear transformation to construct instances 
% similar to the scenario of linear functions under uniform contraints. This would  
% result in expected optimization times of $\Theta(n^2)$ 
% for the \ea. 
%
%

\section{Upper Bound}
\label{sec4}

The following theorem is the main result of this paper, showing that the 
\ea can optimize the generalized class of functions in asymptotically 
the same time as an ordinary linear function.

\begin{theorem}
\label{theo:sumoflinear} 
Let $f$ be the sum of two transformed linear functions as defined in the 
set-up in Section~\ref{sec:two}. Then the expected optimization time of the \ea on $f$ is 
$O(n\log n)$.
\end{theorem}

The proof of Theorem~\ref{theo:sumoflinear} uses drift analysis with a carefully defined potential function, explained in the following.

\paragraph{Potential function.} We build upon the approach from \cite{WittCPC13} to construct a potential 
function~$g^{(1)}$ for $\ell_1$ and a potential 
function $g^{(2)}$ for $\ell_2$, resulting in a combined 
potential function $\phi(x)=g^{(1)}(x)+g^{(2)}(x)$. The individual potential functions are obtained 
in the same way as if the \ea with mutation probability $1/n$ was 
only optimizing  $\ell_1$ and~$\ell_2$, respectively, on an $\alpha n$-dimensional 
and $(1-\alpha)n$-dimensional search space, respectively. The key idea 
is that accepted steps of the \ea on $g$ 
must improve at least one of the two 
functions $\ell_1$ and~$\ell_2$. This event 
leads to a high enough drift of the respective 
potential function that is still positive 
after pessimistically incorporating 
the potential loss due to flipping zero-bits 
that only the other linear function depends on.

We proceed with the definition of the potential functions $g^{(1)}$ and 
$g^{(2)}$ (similarly to Section~5 in \cite{WittCPC13}). For the two underlying 
linear functions we assume their arguments  are reordered according to increasing weights. Note we cannot necessarily sort the set of all indices $1,\dots,n-s$ of the function $f$ so that both underlying linear functions have increasing coefficients; however, as we analyze the underlying functions separately, we can each time use the required sorting in these separate considerations.

\begin{definition}
\label{def:potential}
 Given a linear function $\sum_{i=1}^k w_i x_i$, where $w_1\le \dots\le w_k$, 
  we  define the potential function $g(x_1,\dots,x_k)=\sum_{i=1}^k g_i x_i$ by
 \[
 g_{i} = \left(1+\frac{1}{n}\right)^{\min\{j\le i\;\mid\; w_j = w_i\}-1}.
 \]
 In our scenario, $g^{(1)}(z)$ is the potential 
 function obtained from applying this construction 
 to the $\alpha n$-dimensional linear function $\ell_1(z)$, and proceeding accordingly with 
 the $(1-\alpha)n$-dimensional function $g^{(2)}(y)$ and $\ell_2(y)$. Finally, we define 
 $\phi(x)=g^{(1)}(z)+g^{(2)}(y)$. 
\end{definition}

% For the proof of Theorem~\ref{theo:sumoflinear}, we shall also need  
% the following lemma, which essentially will be used to  establish
% the non-overlapping case
% $s=0$ as worst case.

% \begin{lemma}
% \label{lem:szero-worstcase}
% For $n$ sufficiently large and $0\le s\le n/2$ it holds that
% \[
% \frac{1}{n-s} \left( \left(1+\frac{1}{n-1}\right)^{\alpha n} - 
% \left(1+\frac{1}{n-1}\right)^{s}\right) \le \frac{1}{n}\left(\left(1+\frac{1}{n-1}\right)^{\alpha n} - 1\right).
% \]
% \end{lemma}

% \begin{proof}
% Writing the inequality with common denominator,  we arrive at the equivalent 
%  inequality
% \[
% \frac{s\cdot\left(1+\frac{1}{n-1}\right)^{\alpha n}-n\cdot \left(1+\frac{1}{n-1}\right)^{s}+n-s}{(n-s)n} \le 0,
%  \]
%  which holds for $s=0$. The derivative of the numerator is
%  \[
%  \left(1+\frac{1}{n-1}\right)^{\alpha n}  - n\cdot\ln(1+1/(n-1)) \left(1+\frac{1}{n-1}\right)^{s} -1.
%  \]
%  Since $\alpha \le \ln 2-\epsilon$, the first term is a constant less than~$2$ (assuming~$n$ large enough) 
%  and the negative term in the middle  
%  at most~$-1+O(1/n)$. Hence, the derivative 
%   is negative for all $s\ge 0$, implying that the inequality is true.
% \end{proof}

We can now give the proof of our main theorem.

\begin{proofof}{Theorem~\ref{theo:sumoflinear}} 
Using the potential function from Definition~\ref{def:potential}, we analyze 
the \ea on~$f$, assume an arbitrary, non-optimal search point~$x_t\in\{0,1\}^{n-s}$ and consider the expected change of~$g$ from time~$t$ 
to time~$t+1$. 
We consider an accepted step where the offspring differs from the parent since this is necessary for $g$ to change. That is,
at least one $1$\nobreakdash-bit flips and $f$  does not grow. Let $A$ be the event that 
an offspring~$x'\neq x_t$ is accepted. For~$A$ to occur, it is necessary  
that at least one 
of the two  functions $\ell_1$ and $\ell_2$ does not grow. Since the two cases 
can be handled in an essentially symmetrically manner (they become perfectly symmetrical for $\alpha=n/2$), 
we only analyze the case that $\ell_2$ does not grow and that 
at least one bit in~$B_2$ is flipped from~$1$ to~$0$.
Hence, we consider exactly the situation 
that the \ea with the linear function $\ell_2$ as $(1-\alpha) n$-bit fitness function produces an 
 offspring that is accepted and different from the parent.

Let $Y_t=g^{(2)}(y_t)$, where $y_t$ is the restriction 
of the search point $x_t$ at time~$t$ to the $(1-\alpha) n$ bits in~$B_2$ that $g^{(2)}$ depends on, assuming the indices of $x_t$ to be reordered with respect to 
increasing coefficients $w^{(2)}_1,\dots,w^{(2)}_{(1-\alpha) n}$. 
To compute the drift of~$g$, we distinguish 
between several cases and events in a way similar to the proof 
of Th.~5.1 in \cite{WittCPC13}. Each of these 
cases first bounds the drift of $Y_t$ sufficiently precisely 
and then adds a pessimistic estimate of the drift 
of $Z_t=g^{(1)}(z_t)$, which corresponds to the other linear function on 
bits from~$B_1$, \ie, the function whose value may grow under the event~$A$.
Note that $g^{(1)}$ 
depends on at least as many bits as~$g^{(2)}$ does.

%, assuming the indices of $z_t$ to be reordered with %respect 
%to increasing coefficients $w^{(1)}_1\dots,w^{(1)}_{(\alpha n}$. 

Since the estimate of the drift of~$Z_t$ is always the same, we present it first. 
Let $\tilde{Z}_{t+1}$ denote the $g^{(1)}$-value of the mutated 
bit string~$x'$ (restricted to the bits in~$B_1$). If $x'$ 
is accepted, then $Z_{t+1}=\tilde{Z}_{t+1}$; otherwise $Z_{t+1}=Z_{t}$.
If we pessimistically assume that each bit in $z_t$ (\ie, the restriction 
of $x_t$ to the bits in~$B_1$) is a zero-bit that can flip to $1$, we obtain the upper bound 
\begin{align}
 \expect{\tilde{Z}_{t+1}-Z_{t}\mid Z_t} & \le \frac{1}{n}\sum_{i=1}^{\alpha n} \left(1+\frac{1}{n}\right)^{i-1} 
= \frac{1}{n} \frac{\left(1+\frac{1}{n}\right)^{\alpha n-1}-1}{1/n} \notag \\
%\notag\\
% & \le \frac{1}{n}\sum_{i=1}^{\alpha n} \left(1+\frac{1}{n-1}\right)^{i-1} 
 & \le  e^{\alpha } - 1 \le e^{\ln (2-\epsilon)}-1 
\le 1-\epsilon,
\label{eq:drift-of-zttilde}
\end{align}
where we use that $\alpha < \ln (2-\epsilon)$ for some constant~$\epsilon>0$. 
Also, since $Z_{t+1}=Z_t$ if the mutation is rejected and we only consider 
flipping zero-bits, we have 
under~$A$ (the event that~$x'$ is accepted) that 
\begin{equation}
\expect{Z_{t+1}-Z_{t}\mid Z_t; A} \le \expect{\tilde{Z}_{t+1}-Z_{t}\mid Z_t} \le 1-\epsilon. 
\label{eq:drift-of-zt}
\end{equation}
Note that 
the estimations \eqref{eq:drift-of-zttilde} and \eqref{eq:drift-of-zt}
include the case that $s$ of the bits in~$z_t$ are shared with the 
input string~$y_t$ of the other linear function $g^{(2)}(y_t)$.

We next conduct the detailed 
drift analysis 
 to 
bound 
$\expect{\phi(x_t)-\phi(x_{t+1})\mid x_t }$, considering certain events necessary for~$A$. 
 Two different cases are 
considered.

\emph{Case 1: at least two one-bits in~$y_t$ flip (event~$S_1$)}. 
Let $\tilde{Y}_{t+1}$ denote the $g^{(2)}$-value of the mutated 
bit string~$x'$, restricted to the bits in~$B_2$, under event~$S_1$ \emph{before selection}. If~$x'$ 
is accepted, then $Y_{t+1}=\tilde{Y}_{t+1}$; otherwise $Y_{t+1}=Y_{t}$. 
Since $g_i\ge 1$ for all~$i$,  
every zero-bit in~$y_t$ flips to one with probability at most~$1/n$, and $(1-\alpha)\le \alpha$,
we can re-use the 
estimations from \eqref{eq:drift-of-zttilde}. 
Bounding the contribution of the 
flipping one-bits from below by~$2$, we  obtain
\begin{align*}
 & \expect{Y_t-\tilde{Y}_{t+1}\mid Y_t; S_1}  \ge 2 - \frac{1}{n}\sum_{i=1}^{(1-\alpha) n} \left(1+\frac{1}{n}\right)^{i-1}   \\
 & \quad
 \ge 2 - \frac{1}{n}\sum_{i=1}^{\alpha n} \left(1+\frac{1}{n}\right)^{i-1} 
 \ge  2 - (1-\epsilon) = 1 + \epsilon.
\end{align*}
Along with \eqref{eq:drift-of-zttilde}, we have 
\begin{align*}
  \expect{\phi(x_t)-\phi(x')\mid x_t ; S_1} 
  &  \ge \expect{Y_t-\tilde{Y}_{t+1}\mid Y_t; S_1} - \expect{Z_t-\tilde{Z}_{t+1}\mid Z_t; S_1} \\
 & \ge 2 - (1-\epsilon) - (1-\epsilon)  > \epsilon .
\end{align*}
%for $n$ large enough.

Since the drift of~$\phi$ is non-negative in 
 Case~$1$,  
we estimate it from below
by~$0$ regardless of whether~$A$ occurs or not and focus only on the event defined in 
the following case.

\emph{Case 2: exactly one one-bit in~$y_t$ flips (event~$S_2$)}. 
Let $i^*$ denote the random index of the flipping one-bit in~$y_t$. Moreover, let the function $\beta(i)=\min\{j\le i\mid w^{(2)}_{j}=w^{(2)}_i\}$ 
denote the smallest index   
at most~$i$  with the same weight as~$w^{(2)}_{i}$, \ie, $\beta(i)-1$ is the largest index of a strictly smaller 
weight; using our assumption that the weights are monotonically increasing with their index. 
If at least one zero-bit having the same or a larger weight than bit~$i^*$ flips, neither $\ell_2$ nor $g_2$ change (because the offspring has the same 
 function value or is rejected); hence, we now,  without 
 loss of generality, only consider the subevents of~$S_2$ 
 where all flipping zero-bits have an index of at most~$\beta(i^*)$. (This reasoning is similar
 to the analysis of \emph{Subcase 2.2.2} in 
 the proof of Th.~5 from \cite{WittCPC13}.) 
 
Redefining notation, let $\tilde{Y}_{t+1}$ denote the $g^{(2)}$-value of the mutated 
bit string~$x'$ (restricted to the bits in~$B_2$) under event~$S_2$ \emph{before selection}. If~$x'$ 
is accepted, then $Y_{t+1}=\tilde{Y}_{t+1}$; otherwise $Y_{t+1}=Y_{t}$. 
Recalling that $A$ is the event that the mutation~$x'$ is accepted, we have by 
the law of total probability
\begin{align*}
\expect{Y_t-Y_{t+1}\mid Y_t; S_2} & = \prob{A\mid S_2}\cdot 
\expect{Y_t-\tilde{Y}_{t+1}\mid Y_t; A\cap S_2} \\
& \ge 
\prob{A\mid S_2}\cdot \expect{Y_t-\tilde{Y}_{t+1}\mid Y_t; S_2},
\end{align*}
where the inequality holds 
since the our  estimation of $\expect{Y_t-\tilde{Y}_{t+1}\mid Y_t; S_2}$ below will consider 
exactly one one-bit to flip and assume all zero-bits to flip independently, even though 
already steps flipping two zero-bits right of 
$\beta(i^*)$ may be rejected. 

Moreover, using the law of total probability and \eqref{eq:drift-of-zt}, 
\begin{equation*}
\expect{Z_{t+1}-Z_{t}\mid Z_t; S_2}  \le \prob{A\mid S_2}\cdot 
\expect{\tilde{Z}_{t+1}-Z_t\mid Z_t; S_2} 
\end{equation*}
and therefore 
\begin{align}
 \expect{\phi(x_t)-\phi(x_{t+1})\mid x_t; S_2}  & = \prob{A\mid S_2}\cdot 
\expect{\phi(x_t)-\phi(x')\mid Y_t; A\cap S_2} \notag\\ 
 & \ge \prob{A\mid S_2} 
\expect{(Y_t - \tilde{Y}_{t+1}) - (\tilde{Z}_{t+1} - Z_t)   \mid x_t; S_2}
\label{eq:tildedriftthree}
\end{align}

It holds that $\prob{A\mid S_2} \ge (1-1/n)^{n-1}\ge e^{-1}$ since the mutation flipping~$i^*$ is 
certainly accepted if no other bits flip. To bound the drift, we use  that 
every zero-bit~$j$ right of $\beta(i^*)$ flips with probability 
$1/n$ and contributes $g^{(2)}_j$ to 
the difference $Y_t-\tilde{Y}_{t+1}$. Moreover, the flip of~$i^*$ contributes the term 
$g^{(2)}_{i^*}$ to the difference. 
Altogether, 
\begin{align*}
&\expect{Y_t-\tilde{Y}_{t+1}\mid Y_t; S_2} 
= \left(1+\frac{1}{n}\right)^{\beta(i^*)-1}  - \frac{1}{n} \sum_{j=1}^{\beta(i^*)-1} 
\left(1+\frac{1}{n}\right)^{j-1} \\ 
& = \left(1+\frac{1}{n}\right)^{\beta(i^*)-1} - \frac{1}{n} \left(\frac{\left(1+\frac{1}{n}\right)^{\beta(i^*)-1}-1}{1/n}\right) 
= 1.
\end{align*}
Combining this with \eqref{eq:drift-of-zt}, 
we have
\[
\expect{\phi(x_t) - \phi(x') \mid x_t; S_2} = \expect{(Y_t - \tilde{Y}_{t+1}) - (\tilde{Z}_{t+1} - Z_t)   \mid x_t; S_2} \ge 
1  - (1-\epsilon) = \epsilon.
\]
Altogether, using \eqref{eq:tildedriftthree} and our lower bound 
$\prob{A\mid S_2} \ge e^{-1}$, we have 
the following lower bound on the drift under~$S_2$: 
\[
\expect{\phi(x_t) - \phi(x_{t+1}) \mid x_t; S_2} \ge 
  e^{-1}\epsilon.
\]

Finally, we compute the total drift considering all possible one-bits that can flip 
under~$S_2$. Let $I$ be the set of one-bits in the whole bit string~$x_t$. Since 
the analysis is analogous 
when considering an index~$i\in I$, we still 
consider the situation 
that the corresponding linear function decreases or stays the same 
if $i\in B_2$, \ie, $i$ belongs to~$y_t$ and remark that an analogous
event~$A'$ with respect to the bits $B_1$ and the 
string $z_t$ can be analyzed in the same way.

 Now, 
for $i\in I$, let $F_i$ denote the event that 
bit~$i$ is the only flipping one-bit in the considered part of the bit string and 
let $F$ be the event that 
exactly one bit from~$I$ flips. We have for 
all~$i\in I$ that 
\[
\expect{\phi(x_t) - \phi(x_{t+1}) \mid x_t; F_i} \ge 
  e^{-1}\epsilon.
\]
and therefore also $\expect{\phi(x_t) - \phi(x_{t+1}) \mid x_t; F} \ge e^{-1}\epsilon$. 
It is sufficient to flip one of the $\card{I}$ one-bits and no other bit 
to have an accepted mutation, which has probability at least 
$(\card{I}/n) (1-1/n)^{n-1} \ge \frac{\card{I}}{en}$. We obtain the unconditional drift
\[
\expect{\phi(x_t) - \phi(x_{t+1}) \mid x_t}  \ge \frac{\card{I}}{en} \expect{\phi(x_t) - \phi(x_{t+1}) \mid x_t; F_i} 
\ge \frac{\card{I}e^{-2}}{n} \epsilon,
\]
recalling that we estimated the drift from below by~$0$ if at least two one-bits flip. To 
conclude the proof, we relate the last bound to~$\phi(x_t)$. Clearly, since 
$g_i\le (1+1/n)^{n-1}\le e$ for all $i\in\{1,\dots,\alpha n\}$ and since 
each one-bit can contribute to both~$g^{(1)}(x_t)$ 
and~$g^{(2)}(y_t)$, 
we have $\phi(x_t) \le 2e\card{I}$ so that 
\[
\expect{\phi(x_t) - \phi(x_{t+1}) \mid x_t} \ge \left(\frac{e^{-3}\epsilon}{2n}\right) \phi(x_t) .
\]
Hence, we have established a multiplicative drift of the potential~$\phi$ with a factor 
of $\delta=(e^{-3}\epsilon)/(2n)$ and we obtain   
 the claimed $O(n\log n)$ bound on the expected optimization time via 
the multiplicative drift theorem (Theorem~\ref{theo:multdrift-upper}), using  $X_0\le n(1+1/(n-1))^n = O(n)$ and 
$\smin=1$. 
\end{proofof}

We remark that the drift factor $(e^{-3}\epsilon)/n$ from the previous proof 
can be improved by constant factors using a more detailed 
case analysis; however, since $\epsilon$ can be arbitrarily small and the final 
bound is in $O$-notation, this does not seem worth the 
effort.

\section{Discussion and Conclusions}
Motivated by studies on separable functions and objective functions for chance constrained problems based on the expected value and variance of solutions, we investigated the quite general setting of the sum of two transformed linear functions and established an $O(n \log n)$ bound for the \ea.

We now would like to point out some topics for further investigations.
Our result from Theorem~\ref{theo:sumoflinear} 
has some limitations. First of all, the domains of the two linear functions may not differ very much 
in size; 
more precisely they must be within a factor of $\alpha/(1-\alpha) \le (\ln(2))/(1-\ln(2)) \approx 2.26$. 
With the current pessimistic assumption that an improving mutation only improves one of the two linear 
functions and simultaneously may flip any bit in the other function to~$1$ without the mutation being rejected, we cannot 
improve this to larger size differences for the domain. For the same reason, the result cannot easily 
be generalized to mutation probabilities $c/n$ for arbitrary constants~$c>0$ as shown for the original case 
of simple linear functions in \cite{WittCPC13}. Although that paper also suggests
a different, more powerful class 
of potential functions to handle high mutation probabilities, it seems difficult to apply these more powerful 
potential functions in the presence of our pessimistic assumptions. With stronger 
conditions on $\alpha$, 
it may be possible to extend the present results to mutation probabilities up to $(1+\epsilon)/(n+s)$ %\frank{$(1+\epsilon)/(n+s)$????}
for a positive constant~$\epsilon$  depending on~$\alpha$. However, it would be more interesting to see whether the $O(n\log n)$ bound would also hold for mutation probability $1/n$ for all $s \geq 1$, which would include  
the function~$g(x)$  from the chance-constrained scenario 
in \eqref{eq:sumofidentityandsquare} for the usual 
mutation probability.

\section*{Acknowledgments}

This work has been supported by the Australian Research Council (ARC) through grant FT200100536 
and by the Independent Research Fund Denmark 
through grant DFF-FNU  8021-00260B.

\bibliographystyle{elsarticle-num} 
\bibliography{references}

\end{document}